\documentclass[10pt,twocolumn,letterpaper]{article}

\usepackage{cvpr}              
\usepackage{algorithm}
\usepackage{listings}
\usepackage{makecell}
\usepackage[table]{xcolor}
\usepackage{xcolor}
\usepackage{multirow}
\usepackage{multicol}
\usepackage{appendix}

\definecolor{cvprblue}{rgb}{0.21,0.49,0.74}
\usepackage[pagebackref,breaklinks,colorlinks,allcolors=cvprblue]{hyperref}

\def\paperID{} 
\def\confName{CVPR}
\def\confYear{2026}

\title{Random Wins All: Rethinking Grouping Strategies for Vision Tokens}

\author{%
  Qihang Fan$^{1, 2}$, Yuang Ai$^{1, 2}$, Huaibo Huang$^{1}$\thanks{Huaibo Huang is the corresponding author.}, Ran He$^{1, 2}$\\
  $^1$MAIS \& NLPR, Institute of Automation, Chinese Academy of Sciences, Beijing, China\\
  $^2$School of Artificial Intelligence, University of Chinese Academy of Sciences, Beijing, China\\
  \texttt{fanqihang.159@gmail.com, shallowdream555@gmail.com, }\\
  \texttt{huaibo.huang@cripac.ia.ac.cn, rhe@nlpr.ia.ac.cn}
}

\begin{document}
\maketitle
\begin{abstract}
Since Transformers are introduced into vision architectures, their quadratic complexity has always been a significant issue that many research efforts aim to address. A representative approach involves grouping tokens, performing self-attention calculations within each group, or pooling the tokens within each group into a single token. To this end, various carefully designed grouping strategies have been proposed to enhance the performance of Vision Transformers. Here, we pose the following questions: \textbf{Are these carefully designed grouping methods truly necessary? Is there a simpler and more unified token grouping method that can replace these diverse methods?} Therefore, we propose the random grouping strategy, which involves a simple and fast random grouping strategy for vision tokens. We validate this approach on multiple baselines, and experiments show that random grouping almost outperforms all other grouping methods. For example, compared to the classic Swin Transformer, our random grouping strategy achieves improvements of \textbf{+1.3}, \textbf{+0.9}, and \textbf{+0.9} across three model sizes. When transferred to downstream tasks, such as object detection, random grouping demonstrates even more pronounced advantages. In response to this phenomenon, we conduct a detailed analysis of the advantages of random grouping from multiple perspectives and identify several crucial elements for the design of grouping strategies: \textbf{positional information}, \textbf{head feature diversity}, \textbf{global receptive field}, and \textbf{fixed grouping pattern}. We demonstrate that as long as these four conditions are met, vision tokens require only an extremely simple grouping strategy to efficiently and effectively handle various visual tasks. We also validate the effectiveness of our proposed random method across multiple modalities, including visual tasks, point cloud processing, and vision-language models. Code will be available at \url{https://github.com/qhfan/random}.
\end{abstract}    
\section{Introduction}
\label{sec:intro}

\begin{figure}
    \centering
    \includegraphics[width=0.99\linewidth]{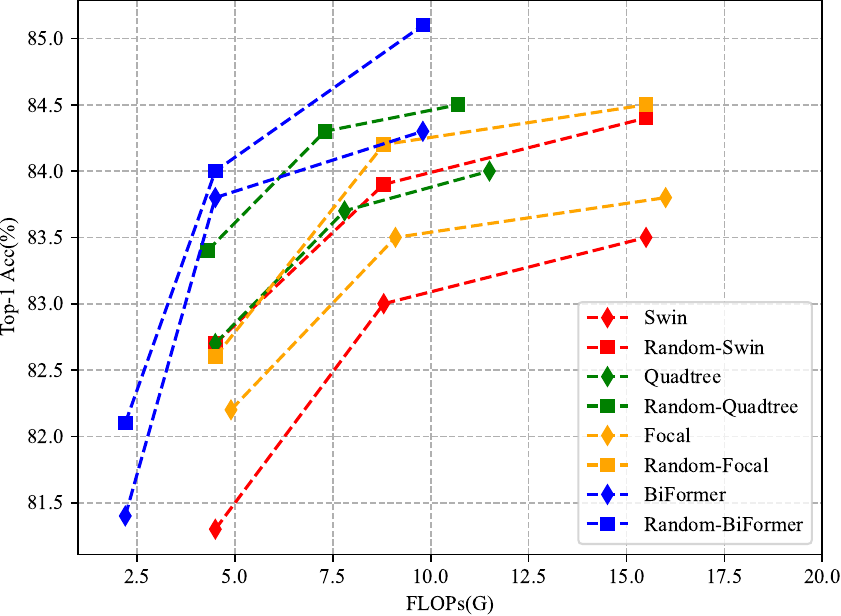}
    \vspace{-3mm}
    \caption{Comparison among baselines and our methods. Random Grouping strategy outperforms all baselines.}
    \vspace{-7mm}
    \label{fig:flops-acc}
\end{figure}

\begin{figure*}[ht]
    \centering
    \includegraphics[width=0.95\linewidth]{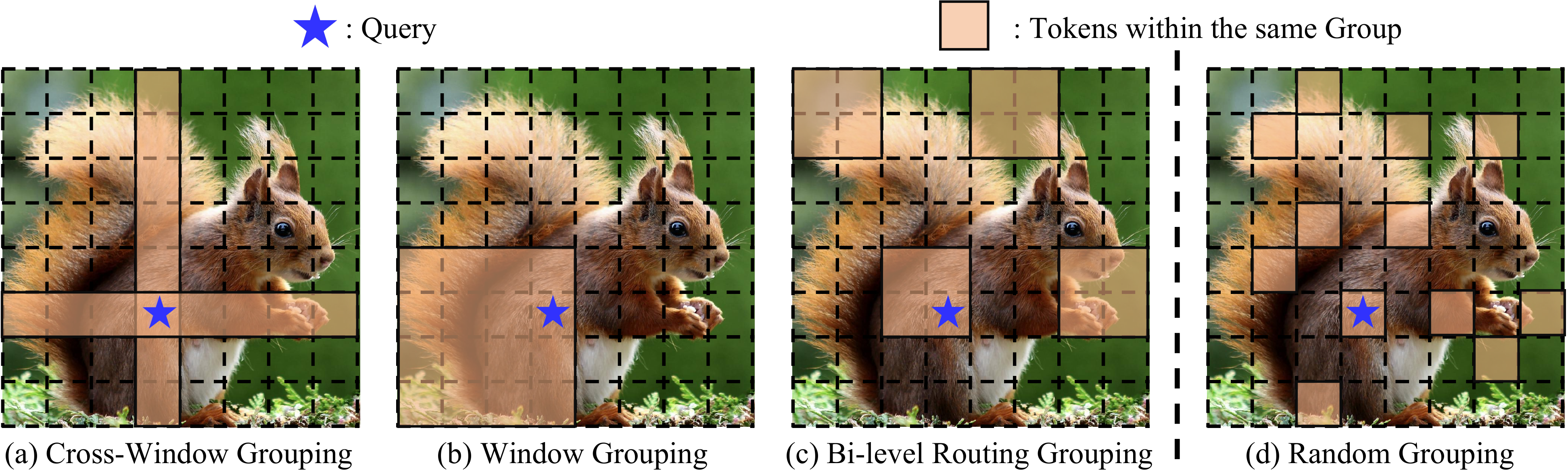}
    \vspace{-3mm}
    \caption{Comparison among different token grouping strategies.}
    \vspace{-6mm}
    \label{fig:method}
\end{figure*}

Transformer, as a powerful architecture, is first introduced in the field of NLP and subsequently finds extensive applications in computer vision~\cite{attention, vit}. It achieves remarkable success in various vision tasks such as image classification, object detection, instance segmentation, and semantic segmentation. 

However, Transformer faces a significant issue: its core operator, self-attention, has quadratic complexity. As the number of vision tokens increases, the computational load of the Vision Transformer grows substantially, greatly limiting its applicability. In recent years, many works attempt to address this issue. Among these efforts, a highly representative approach involves grouping vision tokens. After grouping, self-attention calculations are performed within each group~\cite{SwinTransformer, cswin, fan2024semantic, biformer, crossformer}, or tokens within the same group are fused into a single token, followed by global attention calculations~\cite{pvt, pvtv2, cmt, dat}. Specifically, Swin Transformer~\cite{SwinTransformer} divides all tokens into non-overlapping windows, with self-attention calculations performed within each window. Quadtree~\cite{quadtree}, on the other hand, adopts a context-aware grouping method, using a tree structure to hierarchically group tokens from coarse to fine granularity. BiFormer~\cite{biformer} strikes a balance between the two, combining the context-aware grouping characteristic of Quadtree with the efficiency of grouping seen in Swin Transformer. CrossFormer~\cite{crossformer} not only uses a window-based grouping method but also employs a dilated grouping approach, enabling the model to simultaneously perceive both local and global features.

However, although many carefully designed grouping methods enhance model performance, the complex operations involved in the grouping process significantly impact efficiency and make deployment challenging. This raises the question: are these intricate, meticulously designed grouping methods truly necessary? Is there a simpler and more unified grouping method that can replace these approaches? 

Based on this consideration, we propose an extremely simple token grouping strategy: random grouping strategy, as illustrated in Fig.~\ref{fig:method}. Unlike many methods that require complex grouping operations~\cite{quadtree, focal, sgformer, biformer}, random grouping simply divides all vision tokens into equal segments randomly. Self-attention or pooling is then performed within each group. This random grouping method is not only applicable to basic image classification tasks but also easily transferable to downstream tasks such as object detection and semantic segmentation. Besides, the random group strategy also can be applied to 3D vision tasks such as point cloud segmentation. We experiment with this extremely simple random grouping strategy on various baselines and are surprised to find that this straightforward random approach easily outperforms most of the baselines. Specifically, as shown in Fig.~\ref{fig:flops-acc}, Random-Swin, which employs the random grouping strategy, improves by \textbf{+1.3}, \textbf{+0.9}, and \textbf{+0.9} across different model scales compared to the baseline Swin Transformer. This improvement is even more pronounced in downstream tasks such as object detection. Beyond the visual tasks, the random pattern also can be applied to point cloud task and vision-language models.

In response to this phenomenon, we explain from multiple angles why random grouping achieves such remarkable results. We demonstrate that as long as the four conditions—\textbf{positional information}, \textbf{head feature diversity}, \textbf{global receptive field}, and a \textbf{fixed grouping pattern}—are met, even very simple random grouping methods enable visual models to achieve good results.

Our contributions can be summarized as follows:
\begin{itemize}
    \item We propose a random grouping strategy, an extremely simple and fast grouping method, to reduce the complexity and computational load of Vision Transformers.
    \item We conduct extensive experiments on various baselines and find that this simple strategy not only offers fast inference speed but also outperforms most carefully designed vision token grouping methods.
    \item We analyze the reasons behind the excellent performance of random grouping and identify four key factors that impact the performance of random grouping. We find that as long as these four elements are satisfied, even a simple grouping method can achieve outstanding performance.
\end{itemize}

\section{Related Works}

\paragraph{Vision Transformers.}Transformers originate in the field of NLP and are later introduced to various tasks in computer vision~\cite{attention, vit}. The superior performance of Vision Transformers has garnered significant attention. Many works propose improvements to Vision Transformers; some aim to enhance data efficiency, while others focus on improving computational efficiency~\cite{edgevit, cloformer, MOAT, CaiT, davit, DGT, fan2023rmt, iformer, evit, dynamicvit, dynamicvit2, dualvit2022}. Specifically, DeiT~\cite{deit} incorporates extensive data augmentation~\cite{randomaugment, randera, EMA} into the training process of Vision Transformers and improves data efficiency through model distillation. EViT~\cite{evit} enhances the computational efficiency of Vision Transformers by reducing the number of tokens through the fusion of less important tokens. Aside from these methods, one class of approaches, represented by Swin Transformer~\cite{SwinTransformer}, groups vision tokens based on spatial features or clustering results and performs self-attention calculations within each group. Another class of approaches, represented by PVT~\cite{pvt}, merges vision tokens within the same group into a single token before performing global attention calculations.

\paragraph{Grouping Strategies for Vision Tokens.}As an important approach to addressing the quadratic complexity of Vision Transformers, grouping vision tokens is integral to many Vision Transformer designs. For example, the classic Swin-Transformer~\cite{SwinTransformer} directly groups vision tokens into windows, which limits the receptive field of each token and thereby reduces the complexity of computing self-attention. CSwin-Transformer~\cite{cswin} groups tokens into a cross-window shape, aiming to maximize the receptive field of each token. In addition to these relatively simple and fast methods, there are approaches that use more complex operations for token grouping. For instance, Quadtree~\cite{quadtree} employs a tree structure to hierarchically group tokens, while BiFormer~\cite{biformer} uses a bi-level routing method for token grouping. From the simplest method of partitioning tokens based on their spatial positions~\cite{SwinTransformer, cswin, pvtv2, cmt, litv2, NAT, davit} to more sophisticated approaches that group tokens according to their semantic information~\cite{DGT, quadtree, biformer, dat, fan2024semantic}, model performance improves progressively, but the design of token grouping becomes increasingly complex. To this end, we aim to propose a simple and unified grouping strategy to replace the diverse methods currently in use. Additionally, we seek to identify the key factors that truly contribute to the effectiveness of grouping strategies.

\section{Method}

\begin{figure}[ht]
    \centering
    \includegraphics[width=0.99\linewidth]{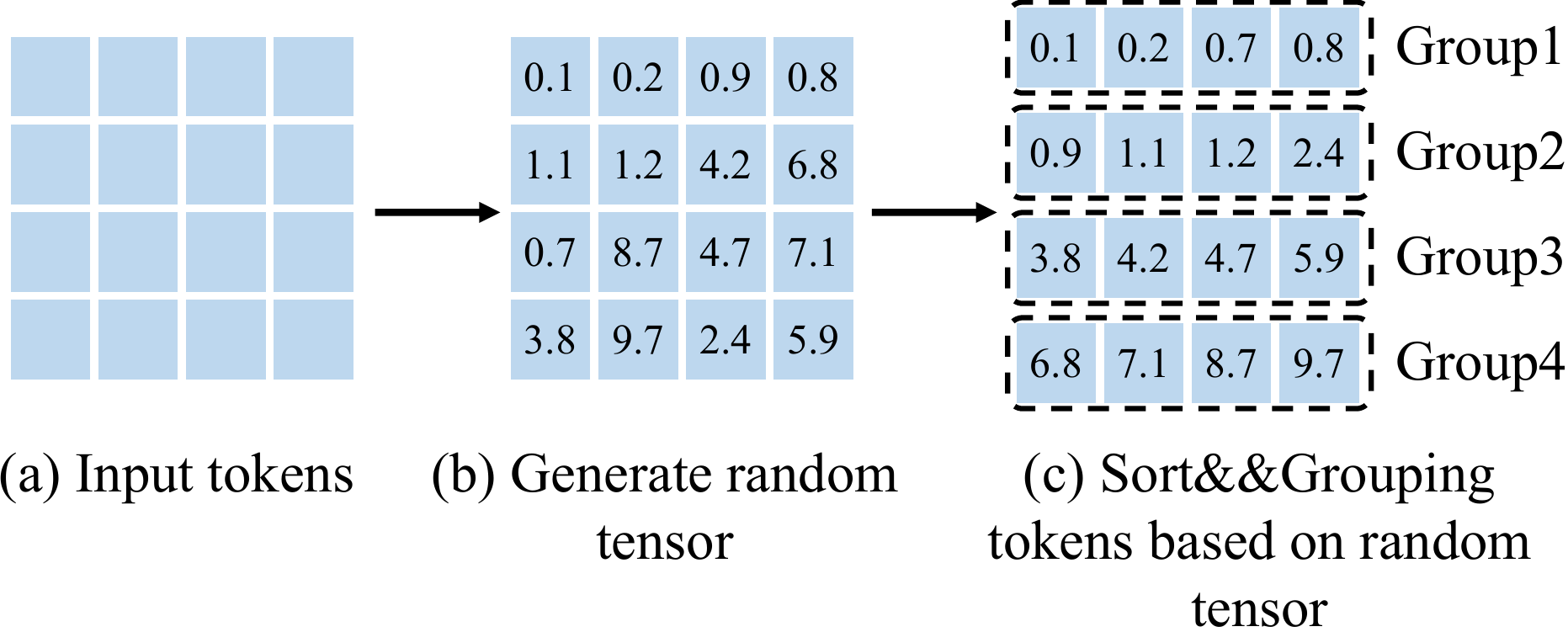}
    \vspace{-4mm}
    \caption{Illustration of the random grouping strategy.}
    \vspace{-4mm}
    \label{fig:randgroup}
\end{figure}

In this section, we explain our proposed random grouping strategy. For simplicity, the subsequent description focuses solely on the single-head scenario.

\paragraph{Generate Random Tensor.} For the input tokens $X\in \mathbb{R}^{h\times w \times d}$, we first generate a random tensor $P\in \mathbb{R}^{h\times w}$ based on the resolution of $X$. The shape of tensor $P$ in the height and width dimensions are the same as those of the input $X$. This means that $P$ corresponds to $X$ on a one-to-one basis, as illustrated in Fig.~\ref{fig:randgroup}. \textbf{Once the tensor $P$ is generated, it is stored. Then, the input tokens of each subsequent image will correspond one-to-one with this stored $P$.} 

\paragraph{Sort Random Tensor.} After obtaining $P$, we sort $P$ in descending order. Since $P$ corresponds to $X$ on a one-to-one basis, $X$ is also sorted in the same order as $P$. 
\textbf{It is worth noting that because $P$ is fixed after it is generated, the input tokens for each image are arranged in the same order.}

\paragraph{Group Vision Tokens.}We denote the rearranged $X$ based on $P$ as $X_p$. By sequentially dividing $X_p$ into equal parts, we obtain the grouped tokens. Since $X_p$ is already randomly shuffled, simply dividing $X_p$ equally will yield the randomly grouped tokens.

\paragraph{Apply $P$ to Higher Resolutions.}Since $P$ is fixed once it is generated and its shape is fixed at $h\times w$, it cannot be directly applied to higher resolution tasks such as object detection, instance segmentation, and semantic segmentation. Therefore, when applying $P$ to higher resolution scenarios, we use nearest-neighbor interpolation to adjust $P$ to match the shape of the input tokens.

\paragraph{Multi-Head Setting.}The above discussion focuses on the single-head scenario. To easily extend this to the multi-head case, we change the shape of the randomly generated tensor $P$ from $h\times w$ to $n\times h\times w$, where $n$ is the number of heads. This means that we use different random tensors to sort the tokens for each head, which results in different random grouping outcomes for each head.

\paragraph{Simplicity and Efficiency.}We did not spend much time describing the random grouping strategy, as it is exceedingly simple. Nevertheless, it outperforms the vast majority of complex grouping methods across many visual tasks, including image classification, object detection, instance segmentation, semantic segmentation, and point cloud segmentation. Additionally, due to its simplicity, it achieves various degrees of acceleration compared to baseline methods. However, the fact that such a seemingly arbitrary method yields excellent results is counterintuitive. Consequently, we dedicate substantial discussion in subsequent sections to attempt to explain this phenomenon.

\paragraph{Baselines.}We categorize the baselines into three types: \textbf{plain backbones}, \textbf{partition-based backbones} that group tokens and perform self-attention within each group, and \textbf{pooling-based backbones} that group tokens and pool them into a single token for further processing. For plain backbones, we apply random grouping and perform self-attention within each group. For the remaining two types, we simply replace their token grouping methods with random grouping. 

\section{Experiments}
To validate the superiority of the random grouping strategy over other methods, we conduct extensive experiments based on multiple baselines, including image classification, object detection, instance segmentation, and semantic segmentation. We also conducted point cloud segmentation based on the point transformer v3~\cite{ptv3}. Besides, we apply random pattern to LLaVA-1.5~\cite{llava1.5} to evaluate the random pattern's effectiveness on vision language model.
\subsection{Image Classification}
\paragraph{Settings.}We train our models on ImageNet-1K~\cite{imagenet} from scratch. We follow the same training strategy in the baselines\cite{deit, SwinTransformer, pvt}, with the only supervision being classification loss for a fair comparison. We use the AdamW optimizer with a cosine decay learning rate scheduler to train the models, and set the initial learning rate, weight decay, and batch size to 0.001, 0.05, and 1024, respectively. We adopt the strong data augmentation and regularization used in baselines~\cite{SwinTransformer, deit, cswin, biformer}, such as RandAugment~\cite{randomaugment} (randm9-mstd0.5-inc1), Mixup~\cite{mixup} (prob=0.8), CutMix~\cite{cutmix} (prob=1.0), and Random Erasing~\cite{randera} (prob=0.25). 

\vspace{-3mm}

\begin{table}[th]
    \centering
    \setlength{\tabcolsep}{0.25mm}
    \scalebox{0.91}{
    \begin{tabular}{c|c c c|c}
    \toprule[1pt]
         Model & \makecell{Params\\(M)} & \makecell{FLOPs\\(G)} & \makecell{Throughput$\uparrow$\\(imgs/s)} & \makecell{Acc\\(\%)} \\
         \midrule[1pt]
         \multicolumn{5}{c}{Plain Backbone}\\
         DeiT-T~\cite{deit} & 6 & 1.3 & 6433 & 72.2 \\
         \rowcolor{gray!30}Random-DeiT-T & 6 & 1.1 & 6682 & 73.1(\textcolor{red}{+0.9})\\
         DeiT-S~\cite{deit} & 22 & 4.6 & 3122 & 79.8 \\
         \rowcolor{gray!30}Random-DeiT-S & 22 & 4.3 & 3313 & 80.9(\textcolor{red}{+1.1})\\
         DeiT-B~\cite{deit} & 87 & 17.6 & 1226 & 81.8 \\
         \rowcolor{gray!30}Random-DeiT-B & 87 & 17.0 & 1348 & 82.5(\textcolor{red}{+0.7})\\
         \midrule[1pt]
         \multicolumn{5}{c}{Partition-Based Backbone}\\
         Swin-T~\cite{SwinTransformer} & 28 & 4.5 & 1738 & 81.3\\
         \rowcolor{gray!30}Random-Swin-T & 28 & 4.5 & 1866 & 82.7(\textcolor{red}{+1.4})\\
         Swin-S~\cite{SwinTransformer} & 50 & 8.7 & 1186 & 83.0 \\
         \rowcolor{gray!30}Random-Swin-S & 50 & 8.7 & 1248 & 83.9(\textcolor{red}{+0.9})\\
         Swin-B~\cite{SwinTransformer} & 88 & 15.4 & 864 & 83.5 \\
         \rowcolor{gray!30}Random-Swin-B & 88 & 15.4 & 902 & 84.4(\textcolor{red}{+0.9})\\
         \midrule[1pt]
         CSwin-T~\cite{cswin} & 22 & 4.3 & 1621 & 82.7 \\
         \rowcolor{gray!30}Random-CSwin-T & 22 & 4.3 & 1763 & 83.1(\textcolor{red}{+0.4})\\
         CSwin-S~\cite{cswin} & 35 & 6.9 & 1036 & 83.6 \\
         \rowcolor{gray!30}Random-CSwin-S & 35 & 6.9 & 1118 & 84.2(\textcolor{red}{+0.6})\\
         CSwin-B~\cite{cswin} & 78 & 15.0 & 716 & 84.2 \\
         \rowcolor{gray!30}Random-CSwin-B & 78 & 15.0 & 778 & 84.7(\textcolor{red}{+0.5})\\
         \midrule[1pt]
         Quadtree-b2~\cite{quadtree} & 24 & 4.5 & 467 & 82.7 \\
         \rowcolor{gray!30}Random-Quadtree-b2 & 21 & 4.3 & 1926 & 83.4(\textcolor{red}{+0.7})\\
         Quadtree-b3~\cite{quadtree} & 46 & 7.8 & 298 & 83.7 \\
         \rowcolor{gray!30}Random-Quadtree-b3 & 36 & 7.3 & 1204 & 84.3(\textcolor{red}{+0.6})\\
         Quadtree-b4~\cite{quadtree} & 64 & 11.5 & 188 & 84.0 \\
         \rowcolor{gray!30}Random-Quadtree-b4 & 49 & 10.7 & 703 & 84.5(\textcolor{red}{+0.5})\\
         \midrule[1pt]
         BiFormer-T~\cite{biformer} & 13 & 2.2 & 1704 & 81.4 \\
         \rowcolor{gray!30}Random-BiFormer-T & 13 & 2.2 & 1903 & 82.1(\textcolor{red}{+0.7}) \\
         BiFormer-STL~\cite{biformer} & 28 & 4.6 & 1366 & 82.7 \\
         \rowcolor{gray!30}Random-BiFormer-STL & 28 & 4.5 & 1562 & 83.1(\textcolor{red}{+0.4}) \\
         BiFormer-S~\cite{biformer} & 26 & 4.5 & 816 & 83.8 \\
         \rowcolor{gray!30}Random-BiFormer-S & 26 & 4.4 & 1028 & 84.0(\textcolor{red}{+0.2}) \\
         BiFormer-B~\cite{biformer} & 57 & 9.8 & 544 & 84.3 \\
         \rowcolor{gray!30}Random-BiFormer-B & 57 & 9.6 & 667 & 85.1(\textcolor{red}{+0.8}) \\
         \midrule[1pt]
         \multicolumn{5}{c}{Pooling-Based Backbone}\\
         PVTv2-B1~\cite{pvtv2} & 13 & 2.1 & 2908 & 78.7 \\
         \rowcolor{gray!30}Random-PVTv2-B1 & 12 & 2.2 & 2898 & 79.6(\textcolor{red}{+0.9})\\
         PVTv2-B2~\cite{pvtv2} & 25 & 4.0 & 1663 & 82.0 \\
         \rowcolor{gray!30}Random-PVTv2-B2 & 21 & 4.2 & 1678 & 82.7(\textcolor{red}{+0.7})\\
         PVTv2-B3~\cite{pvtv2} & 45 & 6.9 & 1152 & 83.2 \\
         \rowcolor{gray!30}Random-PVTv2-B3 & 36 & 7.2 & 1187 & 83.6(\textcolor{red}{+0.4})\\
         \midrule[1pt]
         Focal-T~\cite{focal} & 29 & 4.9 & 583 & 82.2 \\
         \rowcolor{gray!30}Random-Focal-T & 28 & 4.5 & 1842 & 82.6(\textcolor{red}{+0.4}) \\
         Focal-S~\cite{focal} & 51 & 9.1 & 362 & 83.5 \\
         \rowcolor{gray!30}Random-Focal-S & 50 & 8.7 & 1226 & 84.2(\textcolor{red}{+0.7}) \\
         Focal-B~\cite{focal} & 90 & 16.0 & 248 & 83.8 \\
         \rowcolor{gray!30}Ransom-Focal-B & 88 & 15.5 & 887 & 84.5(\textcolor{red}{+0.7}) \\
    \bottomrule[1pt]
    \end{tabular}}
    \vspace{-3mm}
    \caption{Comparison of random grouping strategy with various baselines. The random grouping strategy not only delivers strong performance but also offers faster inference speed compared to other baselines.}
    \vspace{-3mm}
    \label{tab:classification}
\end{table}

\paragraph{Results.}We show the classification results in the Tab.~\ref{tab:classification}. It is evident that compared to the grouping strategies used in other methods, the random grouping strategy demonstrates clear advantages in both speed and performance. Specifically, on the plain backbone, the random grouping strategy reduces the computational load of DeiT, effectively enhancing its speed and performance. Compared to the various grouping strategies in partition-based models, the random grouping strategy also demonstrates better performance and superior inference speed. Specifically, compared to the strategy used in Quadtree, the random grouping strategy achieves more than a threefold increase in speed and a significant performance improvement. Similar conclusions can be drawn when comparing with pooling-based backbones. 

\subsection{Object Detection and Instance Segmentation}
\paragraph{Settings.}Following baselines, we adopt MMDetection~\cite{mmdetection} to implement RetinaNet~\cite{retinanet} and Mask-RCNN~\cite{maskrcnn} to validate the effect of random grouping on object detection and instance segmentation. We use the commonly used ``$1\times$" (12 training epochs) setting for the RetinaNet and Mask R-CNN. Following baselines~\cite{SwinTransformer, pvtv2, biformer}, during training, images are resized to the shorter side of 800 pixels while the longer side is within 1333 pixels. We adopt the AdamW optimizer with a learning rate of 0.0001 and batch size of 16 to optimize the model. The learning rate declines with the decay rate of 0.1 at the epoch 8 and 11. 

\vspace{-3mm}

\begin{table*}[!ht]
    \setlength{\tabcolsep}{1.15mm}
    \centering
    \scalebox{0.86}{
    \begin{tabular}{c|c c|c c c c c c|c c|c c c c c c}
        \toprule[1pt]
        \multirow{2}{*}{Backbone} & \multirow{2}{*}{\makecell{Params\\(M)}} & \multirow{2}{*}{\makecell{FLOPs\\(G)}} & \multicolumn{6}{c|}{Mask R-CNN $1\times$} & \multirow{2}{*}{\makecell{Params\\(M)}} & \multirow{2}{*}{\makecell{FLOPs\\(G)}} & \multicolumn{6}{c}{RetinaNet $1\times$}\\
         & & & $AP^b$ & $AP^b_{50}$ & $AP^b_{75}$ & $AP^m$ & $AP^m_{50}$ & $AP^m_{75}$ & & & $AP^b$ & $AP^b_{50}$ & $AP^b_{75}$ & $AP^b_S$ & $AP^b_{M}$ & $AP^b_{L}$ \\
        
        \midrule[1pt]
        \multicolumn{17}{c}{Partition-Based Backbone} \\
        Swin-T~\cite{SwinTransformer} & 48 & 267 & 43.7 & 66.6 & 47.7 & 39.8 & 63.3 & 42.7 & 38 & 248 & 41.7 & 63.1 & 44.3 & 27.0 & 45.3 & 54.7 \\
        \rowcolor{gray!30}Random-Swin-T & 48 & 267 & \textbf{46.0} & \textbf{68.1} & \textbf{50.5} & \textbf{41.9} & \textbf{65.3} & \textbf{45.4} & 38 & 248 & \textbf{44.3} & \textbf{65.8} & \textbf{46.9} & \textbf{29.7} & \textbf{47.9} & \textbf{57.8} \\
        Swin-S~\cite{SwinTransformer} & 69 & 359 & 45.7 & 67.9 & 50.4 & 41.1 & 64.9 & 44.2 & 60 & 339 & 44.5 & 66.1 & 47.4 & 29.8 & 48.5 & 59.1 \\
        \rowcolor{gray!30}Random-Swin-S & 69 & 359 & \textbf{48.0} & \textbf{69.5} & \textbf{51.9} & \textbf{43.2} & \textbf{67.1} & \textbf{47.0} & 60 & 339 & \textbf{46.6} & \textbf{67.8} & \textbf{48.9} & \textbf{30.5} & \textbf{49.3} & \textbf{62.3} \\
        Swin-B~\cite{SwinTransformer} & 107 & 496 & 46.9 & 69.2 & 51.6 & 42.3 & 66.0 & 45.5 & 98 & 477 & 45.0 & 66.4 & 48.3 & 28.4 & 49.1 & 60.6 \\
        \rowcolor{gray!30}Random-Swin-B & 107 & 496 & \textbf{49.1} & \textbf{71.6} & \textbf{64.2} & \textbf{44.6} & \textbf{68.2} & \textbf{47.4} & 98 & 477 & \textbf{47.4} & \textbf{68.8} & \textbf{50.7} & \textbf{31.3} & \textbf{51.6} & \textbf{64.2} \\
        \midrule[1pt]
        CSwin-T~\cite{cswin} & 42 & 279 & 46.7 & 68.6 & 51.3 & 42.2 & 65.6 & 45.4 & -- & -- & -- & -- & -- & -- & -- & -- \\
        \rowcolor{gray!30}Random-CSwin-T & 42 & 279 & \textbf{47.3} & \textbf{69.3} & \textbf{51.6} & \textbf{42.8} & \textbf{66.4} & \textbf{45.9} & -- & -- & -- & -- & -- & -- & -- & -- \\
        CSwin-S~\cite{cswin} & 54 & 342 & 47.9 & 70.1 & 52.6 & 43.2 & 67.1 & 46.2 & -- & -- & -- & -- & -- & -- & -- & -- \\
        \rowcolor{gray!30}Random-CSwin-S & 54 & 342 & \textbf{48.8} & \textbf{71.3} & \textbf{53.4} & \textbf{44.0} & \textbf{67.9} & \textbf{47.3} & -- & -- & -- & -- & -- & -- & -- & -- \\
        \midrule[1pt]
        BiFormer-S~\cite{biformer} & 45 & -- & 47.8 & 69.8 & 52.3 & 43.2 & 66.8 & 46.5 & 35 & -- & 45.9 & 66.9 & 49.4 & 30.2 & 49.6 & 61.7 \\
        \rowcolor{gray!30}Random-BiFormer-S & 45 & 265 & \textbf{48.4} & \textbf{70.6} & \textbf{52.4} & \textbf{43.7} & \textbf{67.2} & \textbf{46.9} & 35 & 246 & \textbf{46.5} & \textbf{67.4} & \textbf{50.0} & \textbf{30.4} & \textbf{50.6} & \textbf{62.3} \\
        \midrule[1pt]
        \multicolumn{17}{c}{Pooling-Based Backbone} \\
        PVTv2-B1~\cite{pvtv2} & 34 & -- & 41.8 & 64.3 & 45.9 & 38.8 & 61.2 & 41.6 & 24 & -- & 41.2 & 61.9 & 43.9 & 25.4 & 44.5 & 54.3 \\
        \rowcolor{gray!30}Random-PVTv2-B1 & 33 & 216 & \textbf{43.0} & \textbf{66.1} & \textbf{46.9} & \textbf{39.1} & \textbf{52.4} & \textbf{42.9} & 23 & 197 & \textbf{42.4} & \textbf{63.0} & \textbf{44.8} & \textbf{26.6} & \textbf{45.2} & \textbf{57.1} \\
        PVTv2-B2~\cite{pvtv2} & 45 & -- & 45.3 & 67.1 & 49.6 & 41.2 & 64.2 & 44.4 & 35 & -- & 44.6 & 65.6 & 47.6 & 27.4 & 48.8 & 58.6 \\
        \rowcolor{gray!30}Random-PVTv2-B2 & 41 & 252 & \textbf{47.1} & \textbf{68.4} & \textbf{51.0} & \textbf{42.4} & \textbf{65.8} & \textbf{45.6} & 31 & 233 & \textbf{46.0} & \textbf{66.6} & \textbf{49.2} & \textbf{28.6} & \textbf{50.0} & \textbf{59.9} \\
        PVTv2-B3~\cite{pvtv2} & 65 & -- & 47.0 & 68.1 & 51.7 & 42.5 & 65.7 & 45.7 & 55 & -- & 45.9 & 66.8 & 49.3 & 28.6 & 49.8 & 61.4 \\
        \rowcolor{gray!30}Random-PVTv2-B3 & 56 & 342 & \textbf{47.9} & \textbf{69.3} & \textbf{53.0} & \textbf{43.5} & \textbf{66.2} & \textbf{47.3} & 46 & 323 & \textbf{47.0} & \textbf{68.1} & \textbf{49.9} & \textbf{29.8} & \textbf{50.6} & \textbf{62.5} \\
        \bottomrule[1pt]
    \end{tabular}}
    \vspace{-3mm}
    \caption{Comparison of random grouping strategy with various baselines using RetinaNet and Mask R-CNN on COCO val2017 object detection and instance segmentation. }
    \vspace{-7mm}
    \label{tab:COCO1x}
\end{table*}

\paragraph{Results.}We show the comparison results in the Tab.~\ref{tab:COCO1x}. From these results, it is evident that our proposed random grouping strategy achieves improvements in both object detection and instance segmentation, whether using a partition-based backbone or a pooling-based backbone. Specifically, compared to CSwin-S, Random-CSwin-S achieves an increase of +0.9 in $AP^b$ and +0.8 in $AP^m$. The improvements are even more pronounced when compared to the pooling-based PVTv2. 

\subsection{Semantic Segmentation}

\paragraph{Settings.}We adopt the Semantic FPN~\cite{semanticfpn} and UperNet~\cite{upernet} based on MMSegmentation~\cite{mmsegmentation} to validate the superior of random grouping strategy. Following the previous works~\cite{fan2023rmt, hornet, Ortho}, we train the model for 80k iterations with the framework of Semantic FPN, and train the models for 160k iterations with the framework of UperNet.  All models are trained with the input resolution of $512\times 512$. When testing the model, we resize the shorter side of the image to $512$ pixels.

\begin{table}[ht]
    \centering
    \setlength{\tabcolsep}{0.45mm}
    \scalebox{0.85}{
    \begin{tabular}{c|c c c|c c c}
    \toprule[1pt]
    \multirow{3}{*}{Model} & \multicolumn{3}{c|}{Semantic FPN 80K} & \multicolumn{3}{c}{Upernet 160K} \\
    & \makecell{Params\\(M)} & \makecell{FLOPs\\(G)} & \makecell{mIoU\\(\%)} & \makecell{Params\\(M)} & \makecell{FLOPs\\(G)} & \makecell{mIoU$_{ss}$\\(\%)} \\
    \midrule[0.5pt]
    \multicolumn{7}{c}{Plain Backbone} \\
    DeiT-S & -- & -- & -- & 52 & 1099 & 43.0 \\
    \rowcolor{gray!30}Random-DeiT-S & -- & -- & -- & 52 & 1099 & \textbf{44.5} \\
    \midrule[0.5pt]
    \multicolumn{7}{c}{Partition-Based Backbone} \\
    Swin-T~\cite{SwinTransformer} & -- & -- & -- & 60 & 945 & 44.5 \\
    \rowcolor{gray!30}Random-Swin-T & -- & -- & -- & 60 & 945 & \textbf{46.8} \\ 
    Swin-S~\cite{SwinTransformer} & -- & -- & -- & 81 & 1038 & 47.6 \\
    \rowcolor{gray!30}Random-Swin-S & -- & -- & -- & 81 & 1038 & \textbf{48.9} \\
    \midrule[1pt]
    CSwin-T~\cite{cswin} & 26 & 202 & 48.2 & 60 & 959 & 49.3 \\
    \rowcolor{gray!30}Random-CSwin-T & 26 & 202 & \textbf{48.9} & 60 & 959 & \textbf{49.7} \\
    CSwin-S~\cite{cswin} & 39 & 271 & 49.2 & 65 & 1027 & 50.4 \\
    \rowcolor{gray!30}Random-Cswin-S & 39 & 271 & \textbf{49.9} & 65 & 1027 & \textbf{51.1} \\
    CSwin-B~\cite{cswin} & 81 & 464 & 49.9 & 109 & 1222 & 51.1 \\
    \rowcolor{gray!30}Random-Cswin-B & 81 & 464 & \textbf{50.8} & 109 & 1222 & \textbf{52.2} \\
    \midrule[1pt]
    
    BiFormer-S~\cite{biformer} & 29 & -- & 48.9 & 55 & -- & 49.8 \\
    \rowcolor{gray!30}Random-BiFormer-S & 29 & 182 & \textbf{49.7} & 55 & 935 & \textbf{50.4} \\
    BiFormer-B~\cite{biformer} & 60 & -- & 49.9 & 86 & -- & 51.0 \\ 
    \rowcolor{gray!30}Random-BiFormer-B & 60 & 284 & \textbf{51.0} & 86 & 1046 & \textbf{52.0} \\
    \midrule[1pt]
    \multicolumn{7}{c}{Pooling-Based Backbone} \\
    PVTv2-B1~\cite{pvtv2} & 18 & 34 & 42.5 & -- & -- & -- \\
    \rowcolor{gray!30}Random-PVTv2-B1 & 17 & 35 & \textbf{43.6} & -- & -- & -- \\
    PVTv2-B2~\cite{pvtv2} & 29 & 46 & 45.2 & -- & -- & -- \\
    \rowcolor{gray!30}Random-PVTv2-B2 & 13 & 48 & \textbf{46.3} & -- & -- & -- \\
    PVTv2-B3~\cite{pvtv2} & 49 & 62 & 47.3 & -- & -- & -- \\
    \rowcolor{gray!30}Random-PVTv2-B3 & 40 & 65 & \textbf{48.8} & -- & -- & -- \\
    \midrule[1pt]
    Focal-T~\cite{focal} & -- & -- & -- & 62 & 998 & 45.8 \\
    \rowcolor{gray!30}Random-Focal-T & -- & -- & -- & 60 & 945 & \textbf{46.9} \\
    Focal-S~\cite{focal} & -- & -- & -- & 85 & 1130 & 48.0 \\
    \rowcolor{gray!30}Random-Focal-S & -- & -- & -- & 81 & 1038 & \textbf{49.2} \\
    \bottomrule[1pt]
    \end{tabular}}
    \vspace{-3mm}
    \caption{Comparison of random grouping strategy with various baselines using Semantic FPN and UperNet.}
    \vspace{-3mm}
    \label{tab:seg}
\end{table}

\paragraph{Results.}We show the results of semantic segmentation in the Tab.~\ref{tab:seg}. Our simple and fast random grouping method achieves better results compared to various advanced grouping methods. Specifically, under the Semantic FPN framework, using random grouping can achieve a \textbf{+1.1} mIoU improvement in the base model size compared to the relatively more complex Bi-level routing grouping method in BiFormer.

\subsection{Point Cloud Segmentation}
\paragraph{Settings.}We choose the point cloud segmentation task to further validate the superiority of random grouping. Specifically, we select Point Transformer v3~\cite{ptv3}, a powerful point cloud backbone, as our baseline. We train the model on ScanNet~\cite{scannet} and validate it on the validation set. Latency is tested on an A100 GPU.
\begin{table}[ht]
    \centering
    \begin{tabular}{c|c c|c}
    \toprule[1pt]
         Model & \makecell{Params\\(M)} & \makecell{Latency$\downarrow$\\(ms)} & \makecell{mIoU\\(\%)} \\
         \midrule[0.5pt]
         PTv3~\cite{ptv3} & 46 & 88 & 77.6 \\
         \rowcolor{gray!30}Random-PTv3~\cite{ptv3} & 46 & \textbf{68} & \textbf{77.8} \\
    \bottomrule[1pt]
    \end{tabular}
    \vspace{-3mm}
    \caption{Comparison of random grouping strategy and baseline on point cloud segmentation.}
    \vspace{-3mm}
    \label{tab:3dseg}
\end{table}
\paragraph{Results.}As shown in Tab.~\ref{tab:3dseg}, because our proposed random grouping strategy is simpler and easier to operate compared to the point cloud grouping strategy used in Point Transformer v3, it results in lower inference latency than Point Transformer v3. Additionally, random grouping brings a certain degree of performance improvement to the model.

\subsection{Vision Language Models}

\paragraph{Settings.}We follow the setting of LLaVA-1.5~\cite{llava1.5} and LLaVA-1.6 to evaluate the effectiveness of random pattern on the vision-language model. At each Transformer block of the LLM, we permute the input order of vision tokens based on a randomly generated tensor. The positional encodings are aligned with the positions of the vision tokens, while the order of text tokens remains unchanged. The model is trained and evaluated using this strategy.
\begin{table}[ht]
    \centering
    \setlength{\tabcolsep}{0.7mm}
    \scalebox{0.85}{
    \begin{tabular}{c|cccccc}
    \toprule[1pt]
         Model & VQA$^T$ & MME$^P$ & GQA & POPE & VQAv2 & SEED$^I$\\
         \midrule[0.5pt]
         LLaVA-1.5 & 58.2 & 1511 & 62.0 & 86.1 & 78.5 & 66.1 \\
         +Random Pattern & \textbf{58.6} & \textbf{1578} & \textbf{62.3} & \textbf{87.4} & \textbf{78.5} & \textbf{66.9} \\
         \midrule[0.5pt]
         LLaVA-1.6 & 64.9 & 1519 & 64.2 & 86.5 & 81.8 & 70.2 \\
         +Random Pattern & \textbf{65.2} & \textbf{1593} & \textbf{64.5} & \textbf{87.6} & \textbf{82.1} & \textbf{70.6} \\
    \bottomrule[1pt]
    \end{tabular}}
    \vspace{-3mm}
    \caption{Appliying random pattern to LLaVA-1.5/1.6.}
    \vspace{-3mm}
    \label{tab:llava}
\end{table}
\paragraph{Results.}We show the results in the Tab.~\ref{tab:llava}. Applying random pattern to the vision tokens improve the model's performance on all benchmarks.


\subsection{Why Is Random Grouping Better?}
Intuitively, it seems unlikely that a simple random grouping strategy could perform better than other carefully designed, complex methods. Moreover, random grouping causes the model to lose many local biases, making this phenomenon even more intriguing. To explain this, we analyze several aspects to understand why such an extremely simple strategy achieves such good results. Our experiments reveal that as long as a few key elements are satisfied, the specific way tokens are grouped becomes less important. Even with an extremely simple method like random grouping, the model can learn to acquire strong visual representations.

\paragraph{Position Information.}When constructing Random-Swin, the token grouping method is completely random, making it impossible to use the original RPE utilized by Swin. Therefore, we replace RPE with CPE~\cite{CPVT}. To ensure a fair comparison, we train Swin-T using CPE as the positional encoding. We present the results in Tab.~\ref{tab:abpos}. As clearly shown, positional information has a significant impact on random grouping. Unlike window-based grouping, random grouping does not introduce local biases, making positional information particularly crucial. When positional information is added, the performance of the random grouping strategy improves significantly. This indicates that positional information is essential for grouping methods that do not introduce inductive biases. 
\begin{table}[ht]
    \centering
    \setlength{\tabcolsep}{1.1mm}
    \begin{tabular}{c|c c| c}
    \toprule[1pt]
         Model & Params(M) & FLOPs(G) & Acc(\%)\\
         \midrule[0.5pt]
         Swin-T~\cite{SwinTransformer} & 28 & 4.5 & 81.3 \\
         RPE$\xrightarrow{}$CPE & 28 & 4.5 & 81.6 \\
         CPE$\xrightarrow{}$APE & 28 & 4.5 & 80.5 \\
         CPE$\xrightarrow{}$LePE & 28 & 4.5 & 81.7 \\
         w/o PE & 28 & 4.5 & 80.1(\textcolor{red}{-1.6}) \\
         \midrule[0.5pt]
         Random-Swin-T & 28 & 4.5 & 82.7 \\
         CPE$\xrightarrow{}$APE & 28 & 4.5 & 82.3 \\
         CPE$\xrightarrow{}$LePE & 28 & 4.5 & 83.0 \\
         w/o PE & 28 & 4.5 & 79.3(\textcolor{red}{-3.7}) \\
    \bottomrule[1pt]
    \end{tabular}
    \vspace{-3mm}
    \caption{Effect of position information.}
    \vspace{-5mm}
    \label{tab:abpos}
\end{table}

\begin{figure}
    \centering
    \includegraphics[width=0.99\linewidth]{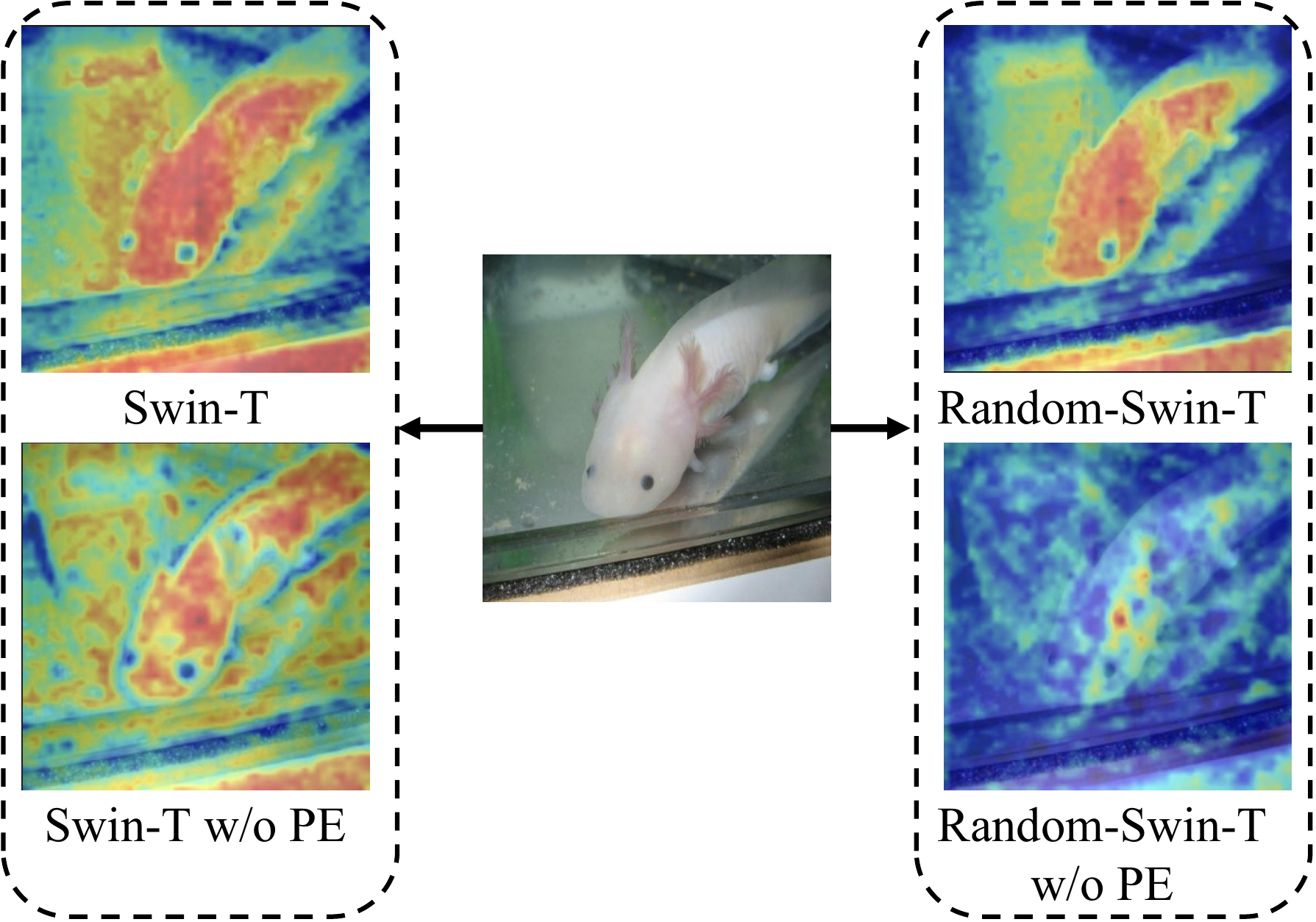}
    \vspace{-3mm}
    \caption{Comparison of attention maps with and without positional information. Positional information is particularly crucial for the random grouping method, which lacks local biases.}
    \vspace{-3mm}
    \label{fig:pemap}
\end{figure}

To further explore the role of positional information in learning visual features, we qualitatively visualize the attention maps learned by different models. The results are shown in Fig.~\ref{fig:pemap}. From the feature maps, it is clear that with positional information, the features learned by Random-Swin are less noisy and more distinct compared to those learned by Swin. However, when positional information is removed, the features learned by both degrade, with Random-Swin experiencing a more severe degradation. Consequently, its performance declines more significantly. This implies that as long as positional information is present, the choice of token grouping strategy becomes less critical.

\paragraph{Head Feature Diversity.}For two heads in multi-head attention, $m$ and $n$, we calculate the cosine similarity between the tokens at all corresponding positions in the two heads, then take the average to obtain the similarity between the feature maps of the two heads. As shown specifically in Eq.~\ref{eq:simhead}:
\begin{equation}
\label{eq:simhead}
    {\rm Sim}(X_m, X_n)=\frac{\sum_{i=1}^N{\rm cos}(X_{mi}, X_{ni})}{N}
\end{equation}
where $X$ is the feature map. $m$, $n$ are the indices of the heads in multi-head attention. $i$ is the index of the token and $N$ is the total number of the token. ${\rm cos}$ denotes the cosine similarity. The value of ${\rm Sim}(X_m, X_n)$ ranges from 0 to 1, with higher values indicating greater similarity between the two heads.

\begin{figure}[ht]
    \centering
    \includegraphics[width=0.99\linewidth]{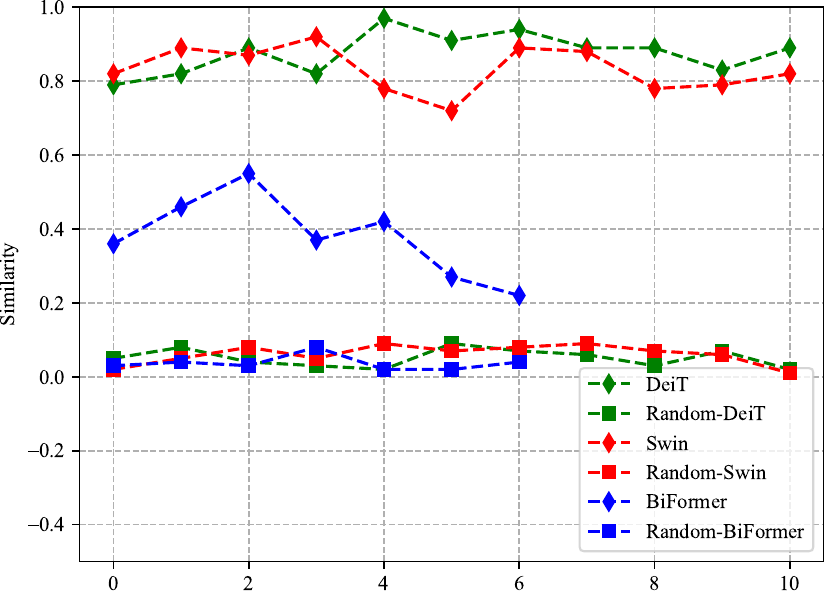}
    \vspace{-3mm}
    \caption{Similarity curves between adjacent heads in different Models}
    \vspace{-3mm}
    \label{fig:head}
\end{figure}

Based on the DeiT-S and the third stage of Swin-T~\cite{SwinTransformer}, BiFormer-S~\cite{biformer}, we visualize the similarity curves between adjacent heads in the feature maps generated by multi-head attention. As shown in Fig.~\ref{fig:head}, compared to the original model, the models using the random grouping strategy exhibit significantly lower similarity between different heads. This reduction occurs because, in random grouping, each head is associated with a different random tensor, leading to distinct grouping methods for each head and, consequently, significantly different learned features between the heads.

\begin{figure}[ht]
    \centering
    \includegraphics[width=0.99\linewidth]{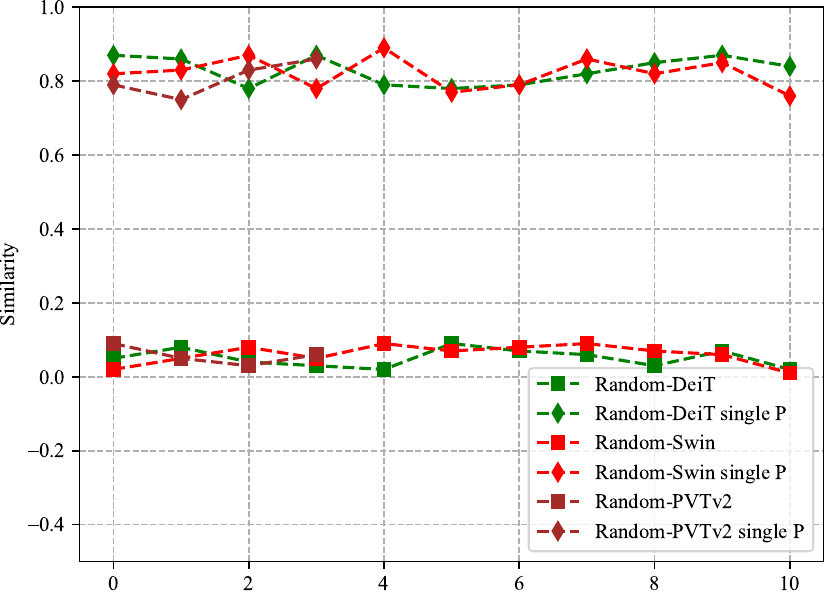}
    \vspace{-3mm}
    \caption{Comparison of each head with a unique random tensor and all heads sharing a single random tensor. Sharing a single random tensor ($P$) significantly reduces the diversity between different heads.}
    \label{fig:head2}
\end{figure}

To validate the impact of head feature diversity on model performance, we modify the original random grouping strategy by using a single random tensor $P$ for all heads instead of a different random tensor for each head. This means all heads follow the same random grouping strategy. As shown in Fig.~\ref{fig:head2}, this modification significantly increases the similarity between the heads, indicating a reduction in head feature diversity.
\begin{table}[ht]
    \centering
    \setlength{\tabcolsep}{0.1mm}
    \scalebox{0.9}{
    \begin{tabular}{c|c c|c}
        \toprule[1pt]
        Model & Params(M) & FLOPs(G) & Acc(\%) \\
        \midrule[0.5pt]
        DeiT-S~\cite{deit} & 22 & 4.6 & 79.8 \\
        Random-DeiT-S & 22 & 4.3 & 80.9 \\
        multi $P$$\xrightarrow{}$single $P$ & 22 & 4.3 & 79.2(\textcolor{red}{-1.7}) \\
        \midrule[0.5pt]
        Swin-T~\cite{SwinTransformer} & 28 & 4.5 & 81.3 \\
        Random-Swin-T & 28 & 4.5 & 82.7 \\
        multi $P$$\xrightarrow{}$single $P$ & 28 & 4.5 & 80.5(\textcolor{red}{-2.2}) \\
        \midrule[0.5pt]
        BiFormer-STL~\cite{biformer} & 28 & 4.6 & 82.7 \\
        Random-BiFormer-STL~\cite{biformer} & 28 & 4.5 & 83.1 \\
        multi $P$$\xrightarrow{}$single $P$ & 28 & 4.5 & 80.8(\textcolor{red}{-2.3}) \\
        \midrule[0.5pt]
        PVTv2-B1~\cite{pvtv2} & 13 & 2.1 & 78.7 \\
        Random-PVTv2-B1 & 12 & 2.2 & 79.6 \\
        multi $P$$\xrightarrow{}$single $P$ & 12 & 2.2 & 78.1(\textcolor{red}{-1.5}) \\
        \bottomrule[1pt]
    \end{tabular}}
    \vspace{-3mm}
    \caption{Effect of head feature diversity.}
    \vspace{-5mm}
    \label{tab:headdiv}
\end{table}
As shown in Tab.~\ref{tab:headdiv}, we provide a detailed comparison of the impact of head feature diversity on model performance. It is evident that when all heads use the same random strategy, resulting in reduced head feature diversity, the performance of various models significantly declines. This clearly demonstrates the crucial role that head feature diversity plays in learning robust visual features. Additionally, the performance of Random-BiFormer-STL illustrates this point. Since BiFormer-STL inherently has lower feature similarity between heads, indicating greater feature diversity (as shown in Fig.~\ref{fig:head}), the increase in head feature diversity when using the random grouping strategy is less pronounced, resulting in a smaller performance improvement.

\paragraph{Global Receptive Field.}The global receptive field is a significant advantage of Vision Transformers over architectures like CNNs~\cite{resnet, convnext}. However, most partition-based methods compromise this by reducing the global receptive field to decrease the model's computational load~\cite{SwinTransformer, cswin, dat, LVT}. Unlike previous grouping methods, random grouping still manages to sparsely capture global information. This is likely one of the reasons for its superior performance. 

To investigate the impact of the global receptive field on model performance, we enhanced the simplest random grouping strategy. Specifically, we constrain the random numbers corresponding to the tokens in each region within certain intervals. Additionally, these intervals for different regions have a certain degree of overlap. This ensures that, during random grouping, most tokens from the same region remain together, while a portion is randomly swapped to other regions.

\begin{table}[ht]
    \centering
    \setlength{\tabcolsep}{0.9mm}
    \begin{tabular}{c|c c|c}
    \toprule[1pt]
         Model & Params(M) & FLOPs(G) & Acc(\%) \\
         \midrule[0.5pt]
         DeiT-S~\cite{deit} & 22 & 4.6 & 79.8 \\
         Random-DeiT-S & 22 & 4.3 & 80.9 \\
         global$\xrightarrow{}$region & 22 & 4.3 & 79.3(\textcolor{red}{-1.6}) \\
         \midrule[0.5pt]
         Swin-T~\cite{SwinTransformer} & 28 & 4.5 & 81.3 \\
         Random-Swin-T & 28 & 4.5 & 82.7 \\
         global$\xrightarrow{}$region & 28 & 4.5 & 81.5(\textcolor{red}{-1.2}) \\
         \midrule[0.5pt]
         PVTv2-B1~\cite{pvtv2} & 13 & 2.1 & 78.7 \\
         Random-PVTv2-B1 & 12 & 2.2 & 79.6 \\
         global$\xrightarrow{}$region & 12 & 2.2 & 79.3(\textcolor{red}{-0.3}) \\
    \bottomrule[1pt]
    \end{tabular}
    \vspace{-3mm}
    \caption{Effect of global receptive field.}
    \vspace{-3mm}
    \label{tab:abglobal}
\end{table}

As shown in Tab.~\ref{tab:abglobal}, we observe that when the global receptive field is sacrificed to a certain degree, the performance of DeiT and Swin drops significantly. This indicates the crucial role that the global receptive field plays in learning visual features. In contrast, the performance drop for PVTv2 is less pronounced. This is because the pooling-based backbone inherently has a global receptive field. The performance decline in PVTv2 may not be due to the loss of the global receptive field but rather because this region-specific random approach reduces the variety of grouping methods for each head in multi-head attention, thereby decreasing head feature diversity.

\vspace{-3mm}

\paragraph{Fixed Grouping Pattern.}Although we use random grouping, the random tensor $P$ that dictates our grouping method is fixed once generated. Therefore, the random grouping remains the same for different input images. In other words, even though it is random grouping, it follows a consistent pattern just like other methods with specific grouping rules. 

To verify whether the fixed grouping pattern affects the performance of random grouping, we replace it with a fully random approach. For each input image, we use a unique random tensor $P$ to determine the random grouping method. This disrupts the fixed grouping pattern in the original random grouping. By doing so, we explore the influence of the fixed grouping pattern on the performance of random grouping. The results, shown in Tab.~\ref{tab:abfix}, indicate that when using fully random grouping—where the fixed grouping pattern is disrupted—the performance of all models declines significantly. This clearly demonstrates the crucial role that a fixed grouping pattern plays in the effectiveness of the random grouping strategy.

\begin{table}[ht]
    \centering
    \setlength{\tabcolsep}{0.9mm}
    \begin{tabular}{c|c c|c}
    \toprule[1pt]
         Model & Params(M) & FLOPs(G) & Acc(\%) \\
         \midrule[0.5pt]
         DeiT-S~\cite{deit} & 22 & 4.6 & 79.8 \\
         Random-DeiT-S & 22 & 4.3 & 80.9 \\
         Fully Random & 22 & 4.3 & 74.3(\textcolor{red}{-6.6}) \\
         \midrule[0.5pt]
         Swin-T~\cite{SwinTransformer} & 28 & 4.5 & 81.3 \\
         Random-Swin-T~\cite{SwinTransformer} & 28 & 4.5 & 82.7 \\
         Fully Random & 28 & 4.5 & 76.4(\textcolor{red}{-6.3}) \\
         \midrule[0.5pt]
         PVTv2-B1~\cite{pvtv2} & 13 & 2.1 & 78.7 \\
         Random-PVTv2-B1 & 12 & 2.2 & 79.6 \\
         Fully Random & 12 & 2.2 & 75.4(\textcolor{red}{-4.2}) \\
    \bottomrule[1pt]
    \end{tabular}
    \vspace{-3mm}
    \caption{Effect of the fixed grouping pattern.}
    \vspace{-5mm}
    \label{tab:abfix}
\end{table}

\paragraph{Roadmap from Fully Random Grouped Backbone to Random-Swin.}To more clearly illustrate the role of each component, we gradually transform a fully random grouped backbone into Random-Swin-T. By sequentially adding fixed grouping pattern, multi random tensors $P$, and positional encoding into the backbone, the model's performance significantly improves. \textbf{This validates the conclusion that even very simple random grouping can achieve good results, as long as certain conditions are met.}

\begin{table}[ht]
    \centering
    \setlength{\tabcolsep}{0.9mm}
    \scalebox{0.9}{
    \begin{tabular}{c|c c c|c}
    \toprule[1pt]
         Model & \makecell{Params\\(M)} & \makecell{FLOPs\\(G)} & \makecell{Throughput\\(imgs/s)}& \makecell{Acc\\(\%)} \\
         \midrule
         Swin-T & 28 & 4.5 & 1738 & 81.3 \\
         RPE$\xrightarrow{}$CPE & 28 & 4.5 & 1698 & 81.6 \\
         \midrule
         Fully Random(single $P$) & 28 & 4.5 & 1922 & 71.2 \\
         +fixed pattern & 28 & 4.5 & 1922 & 77.6 \\
         +multi $P$ & 28 & 4.5 & 1917 & 80.1 \\
         +CPE & 28 & 4.5 & 1866 & 82.7 \\
         \bottomrule[1pt]
    \end{tabular}}
    \vspace{-4mm}
    \caption{Roadmap from fully random grouping backbone to Random-Swin. Inference speed are measured on A100 GPU.}
    \vspace{-4mm}
    \label{tab:my_label}
\end{table}

\section{Conclusion}
In this paper, we pose the question: Are the complex grouping strategies for vision tokens truly necessary? Is there a simpler and more unified token grouping method that can replace these diverse methods? To address these, we propose an extremely simple random grouping strategy. Experiments show that the random grouping strategy outperforms most complex grouping strategies across various tasks. Moreover, due to its simplicity and ease of implementation, models based on the random grouping strategy achieve faster inference speeds compared to baselines. The success of the random grouping strategy effectively answers our initial questions.
To understand the success of this extremely simple grouping strategy, we conduct an in-depth analysis of the underlying factors: positional information, head feature diversity, global receptive field, and a fixed grouping pattern. Our findings demonstrate that as long as these four elements are present, even the simplest random grouping strategy can achieve good results.
{
    \small
    \bibliographystyle{ieeenat_fullname} 
    \bibliography{main}
}


\end{document}